%% file: main.tex
\documentclass[10pt,twocolumn,letterpaper]{article}

\usepackage[pagenumbers]{cvpr} 

\usepackage{graphicx}
\usepackage{amsmath}
\usepackage{amssymb}
\usepackage{booktabs}
\usepackage{mathtools}
\usepackage{ulem}
\usepackage{subcaption}
\usepackage{multirow}
\DeclarePairedDelimiter\ceil{\lceil}{\rceil}
\DeclarePairedDelimiter\floor{\lfloor}{\rfloor}

\usepackage[pagebackref,breaklinks,colorlinks]{hyperref}

\usepackage[capitalize]{cleveref}
\crefname{section}{Sec.}{Secs.}
\Crefname{section}{Section}{Sections}
\Crefname{table}{Table}{Tables}
\crefname{table}{Tab.}{Tabs.}

\def\cvprPaperID{3502} 
\def\confName{CVPR}
\def\confYear{2023}

\begin{document}

\title{G-CAME: Gaussian-Class Activation Mapping Explainer for Object Detectors}

\author{Quoc Khanh Nguyen\textsuperscript{\rm 1}, Truong Thanh Hung Nguyen\textsuperscript{\rm 1,2},  
Vo Thanh Khang Nguyen\textsuperscript{\rm 1},  \\ Van Binh Truong\textsuperscript{\rm 1}, Quoc Hung Cao\textsuperscript{\rm 1}\\
\textsuperscript{\rm 1}Quy Nhon AI, FPT Software, \textsuperscript{\rm 2}Friedrich-Alexander-Universität Erlangen-Nürnberg\\
{\tt\small \{khanhnq33, hungntt, khangnvt1, binhtv8, hungcq3\}@fsoft.com.vn}
}
\maketitle

\begin{abstract}
    \input{abstract}
\end{abstract}

\section{Introduction}\label{sec:introduction}
\input{introduction}

\section{Related Work}\label{sec:related-work}
\input{related-work}

\section{Methods}\label{sec:proposed-methods}
\input{proposed-methods}

\section{Experiments and Results}\label{sec:results}
\input{results}

\section{Conclusion}\label{sec:Conclusion}
\input{conclusion}


{\small
\bibliographystyle{ieee_fullname}
\bibliography{ref}
}

\section*{Appendix}
\input{appendix}
\end{document}

%% file: abstract.tex
Nowadays, deep neural networks for object detection in images are very prevalent.
However, due to the complexity of these networks, users find it hard to understand why these objects are detected by models.
We proposed Gaussian Class Activation Mapping Explainer (G-CAME), which generates a saliency map as the explanation for object detection models.
G-CAME can be considered a CAM-based method that uses the activation maps of selected layers combined with the Gaussian kernel to highlight the important regions in the image for the predicted box.
Compared with other Region-based methods, G-CAME can transcend time constraints as it takes a very short time to explain an object.
We also evaluated our method qualitatively and quantitatively with YOLOX~\cite{ge2021yolox} on the MS-COCO 2017 dataset~\cite{lin2014microsoft} and guided to apply G-CAME into the two-stage Faster-RCNN~\cite{ren2015faster}) model.

%% file: introduction.tex
In object detection, deep neural networks (DNNs)\cite{girshick2014rich} have significantly improved with the adoption of convolution neural networks.
However, the deeper the network is, the more complex and opaque it is to understand, debug or improve.
To help humans have a deeper understanding of the model's decisions, several eXplainable Artificial Intelligence (XAI) methods using saliency maps to highlight the important regions of input images have been introduced.

One simple and common way to explain the object detector is to ignore the model architecture and only consider the input and output. 
This approach aims to determine the importance of each region in the input image based on the change in the model's output. 
For example, D-RISE\cite{petsiuk2021black}, an improvement of RISE\cite{petsiuk2018rise}, estimates each region's effect on the input image by creating thousands of perturbed images, then feeds them into the model to predict and get the score for each perturbed mask.
Another method is SODEx\cite{sejr2021surrogate}, which is an upgrade of LIME\cite{ribeiro2016should}.
It also uses the same technique as D-RISE to explain object detectors.
In contrast with D-RISE, SODEx gives each super-pixel score in the input image.
Although the results of both SODEx and D-RISE are compelling, the generation of a large number of perturbations slows these methods down considerably.

Other approaches, such as CAM\cite{zhou2016learning} and GradCAM\cite{selvaraju2017grad}, use the activation maps of a specific layer in the model's architecture as the main component to form the explanation.
These methods are faster than mentioned region-based methods but still have some meaningless information since the feature maps are not related to the target object \cite{zhang2021group}. 
Such methods can give a satisfactory result for the classification task. Still, they cannot be applied directly to the object detection task
because these methods highlight all regions having the same target class and fail to focus on one specific region.

In this paper, we propose the \textit{Gaussian Class Activation Mapping Explainer} (G-CAME), which can explain the classification and localization of the target objects. 
Our method improves previous CAM-based XAI methods since it is possibly applied to object detectors. 
By adding the Gaussian kernel as the weight for each pixel in the feature map, G-CAME's final saliency map can explain each specific object. 

Our contributions can be summarized as follows:
\begin{enumerate}
    \item We propose a novel CAM-based method, G-CAME, to explain object detectors as a saliency map. 
    Our method can give an explanation in a reasonably short time, which overcomes the existing methods' time constraints like D-RISE\cite{petsiuk2021black} and SODEx\cite{sejr2021surrogate}.
    \item We propose a simple guide in applying G-CAME to explain two types of commonly used models: YOLOX~\cite{ge2021yolox} (one-stage detector) and Faster-RCNN~\cite{ren2015faster} (two-stage detector).
    \item We qualitatively and quantitatively evaluate our method with D-RISE and prove that our method can give a less noise and more accurate saliency map than D-RISE.
\end{enumerate}

\begin{figure}[]
  \centering
    \includegraphics[scale=0.4]{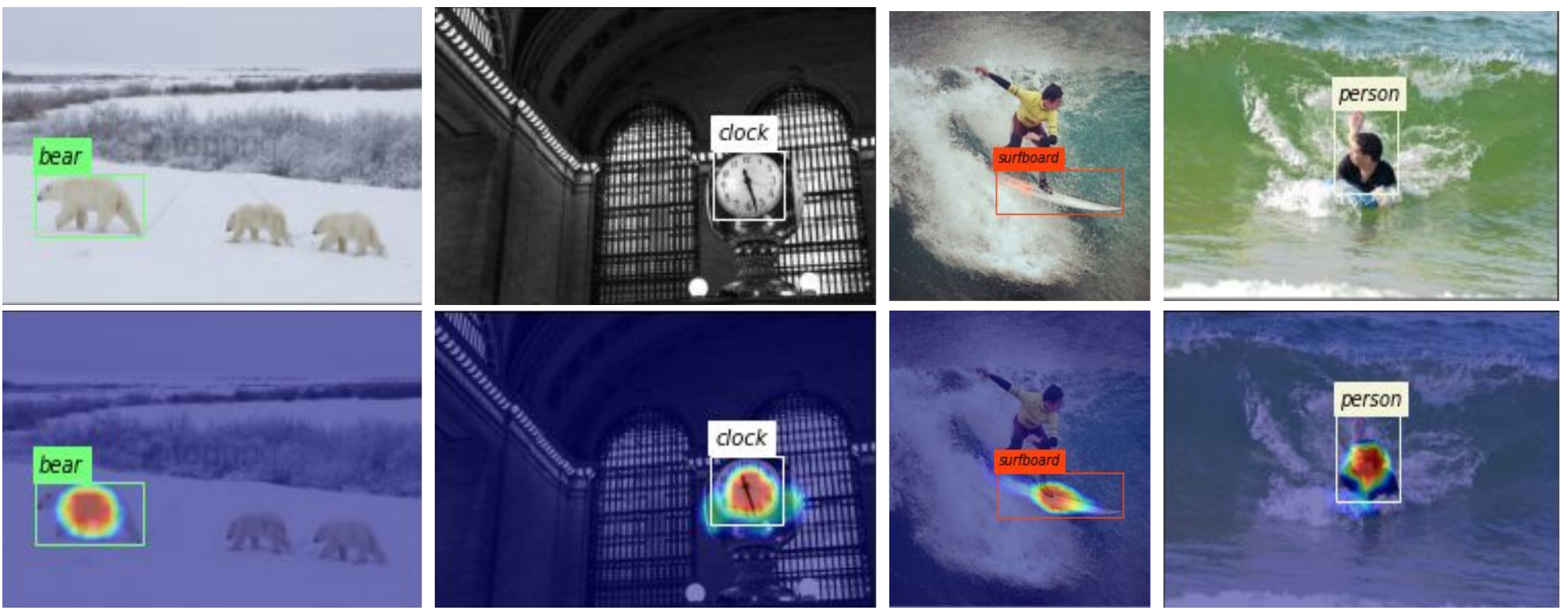}
    \caption{G-CAME can highlight regions that affect the target prediction on sample images in MS-COCO 2017 dataset.}
    \label{fig:my_label}
\end{figure}

%% file: related-work.tex
 \subsection{Object Detection}
Object detection problem is one of the fields in computer vision (CV). 
Object detection models are categorized into two types: one-stage model and two-stage model.
In detail, the one-stage model detects directly over a dense sampling of locations, such as YOLO series\cite{redmon2016you}, SSD\cite{liu2016ssd}, and RentinaNET\cite{lin2017focal}. 
While the two-stage model detects after two phases.
In the first phase, the Region Proposal stage, the model selects a set of Region of Interest (ROI) from the feature extraction stage.
Then, in the second stage, the model classifies based on each proposed ROI. 
Some of the most popular two-stage detection models are R-CNN family~\cite{girshick2014rich}, FPN~\cite{lin2017feature}, and R-FCN~\cite{dai2016r}.

\subsection{Explainable AI}
In CV, several XAI methods are used to analyze deep CNN models in the classification problem, while in the object detection problem, the number of applicable XAI methods is limited.
In general, there are two types of analyzing the model's prediction. 
One is based on the input region to give the saliency map, called \textit{Region-based saliency methods}, while \textit{CAM-based saliency methods} uses feature maps to create the saliency map for the input.
\subsubsection{Region-based saliency methods}
The first type of XAI, Region-based saliency methods, adopt masks to keep a specific region of the input image to measure these regions' effect on output by passing the masked input through the model and calculating each region's weight.
In the classification problem, LIME~\cite{ribeiro2016should} uses several random masks and then weights them by a simple and interpretable model like Linear Regression or Lasso Regression.
An improvement of LIME is RISE\cite{petsiuk2018rise}, in which the authors first generate thousands of masks and employ them to mask the input, then linearly combine them with their corresponding weight score to create the final saliency map.
Several methods are adjusted to apply to the object detection problem.
Surrogate Object Detection Explainer\cite{sejr2021surrogate} (SODEx) employed LIME to explain object detectors.
Instead of calculating the score of each region for the target class like LIME, the author proposed a new metric that calculates the score of each region for the target bounding box.
Detector Randomized Input Sampling for
Explanation (D-RISE)\cite{petsiuk2021black} was proposed as an improvement over RISE.
D-RISE defines a different metric to compute the weighted score for each random mask, then linearly combines them to explain the target bounding box.
All mentioned methods are intuitive since users do not require to understand the model's architecture.
One more common thing is that the explanation is sensitive to the hyperparameters modification.
However, this is also one of the weaknesses of these methods because we can have multiple explanations for one object.
Therefore, if we want a clear and satisfactory explanation, we must choose the hyperparameters carefully.
Another weakness of these methods is taking a lot of time to explain.
\subsubsection{CAM-based methods}
The other approach in XAI is \textit{CAM-based} methods.
In this approach, we must access and explicitly understand the model's architecture.
Class Activation Mapping (CAM)~\cite{zhou2016learning} is the first method to combine the weighted activation map of one or multiple selected convolution layers to form the explanation.
After that, GradCAM~\cite{selvaraju2017grad}, GradCAM++~\cite{chattopadhay2018grad}, and XGradCAM~\cite{fu2020axiom} extended CAM to obtain the saliency map with fine-grained details. 
These methods use the partial derivatives of the selected layers' feature maps concerning the target class score produced by the CNN model to get a weight for each activation map.
CAM-based methods are usually faster than Region-based methods because they only require one or some model's layers to form the explanation and only need to execute a single forward or backward pass. 
However, CAM-based methods' saliency maps usually contain meaningless features and depend entirely on feature maps. 
Also, all previous CAM-based XAI methods are used for the classification problem, but none of them has been proposed for the object detection problem yet. 

In this paper, we proposed G-CAME, a CAM-based method that can explain both one-stage and two-stage object detection models.

%% file: proposed-methods.tex
\begin{figure*}
    \centering
    \includegraphics[scale=0.7]{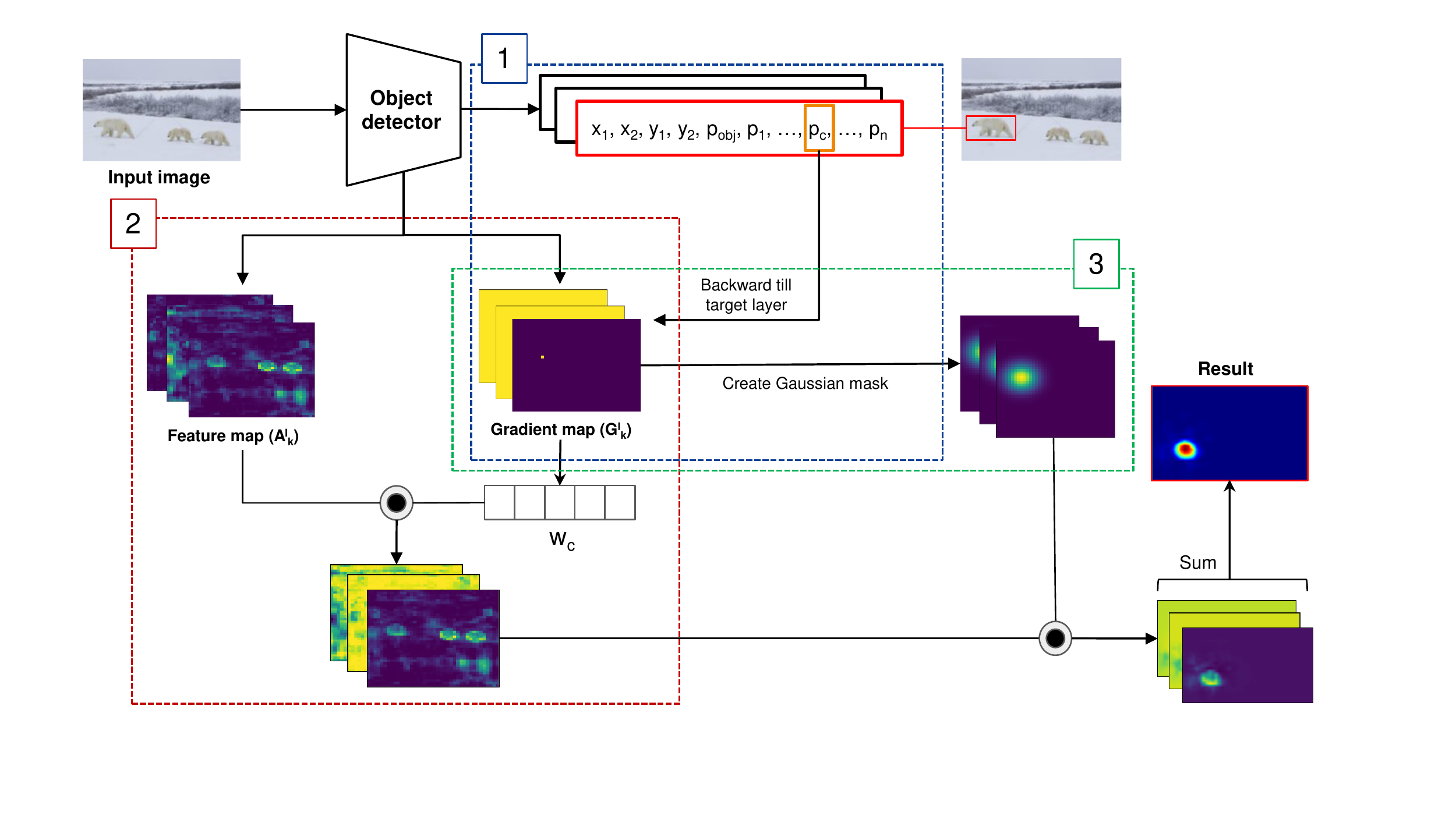}
    \caption{Overview of G-CAME method. We use the gradient-based technique to get the target object's location and weight for each feature map. 
    We multiply element-wise with Gaussian kernel for each weighted feature map to remove unrelated regions. 
    The output saliency map is created by a linear combination of all weighted feature maps after applying the Gaussian kernel.}
    \label{fig:method_overview}
\end{figure*}

For a given image $I$ with size $h$ by $w$, an object detector $f$ and the prediction of that model $d$ includes the bounding box and predicted class. 
We aim to provide a saliency map $S$ to explain why the model has that prediction.
The saliency map $S$ has the same size as the input $I$.
Each value $S_{(i, j)}$ shows the importance of each pixel $(i, j)$ in $I$, respectively, influencing $f$ to give prediction $d$. 
We propose a new method that helps to produce that saliency map in a white-box manner.
Our method is inspired by GradCAM~\cite{selvaraju2017grad}, which uses the class activation mapping technique to generate the explanation for the model's prediction.
The main idea of our method is to use normal distribution combined with the CAM-based method to measure how one region in the input image affects the predicted output. 
Fig.~\ref{fig:method_overview} shows an overview of our method.

We cannot directly apply XAI methods for the classification model to the object detection model because of their output difference.
In the classification task, the model only gives one prediction that shows the image's label. 
However, in the object detection task, the model gives multiple boxes with corresponding labels and the probabilities of objects.

Most object detectors, like YOLO~\cite{redmon2016you} and R-CNN\cite{girshick2014rich}, usually produce $N$ predicted bounding boxes in the format:
\begin{equation}
    d_i = (x^i_1, y^i_1, x^i_2, y^i_2, p_{obj}^i, p^i_1, …, p^i_C) 
\end{equation}
The prediction is encoded as a vector $d_i$ that consists of:
\begin{itemize}
    \item Bounding box information: $(x^i_1, y^i_1, x^i_2, y^i_2)$ denotes the top-left and bottom-right corners of the predicted box.
    \item Objectness probability score: $p_{obj}^i \in [0, 1]$ denotes the probability of an object's occurrence in predicted box.
    \item Class score information: $(p^i_1, …, p^i_C)$ denotes the probability of $C$ classes in predicted box.
\end{itemize}

In almost object detectors, such as Faster-RCNN~\cite{ren2015faster}, YOLOv3~\cite{redmon2018yolov3}, YOLOX~\cite{ge2021yolox}, the anchor boxes technique is widely used to detect bounding boxes.
G-CAME utilizes this technique to find and estimate the region related to the predicted box.
Our method can be divided into 3 phases (Fig.~\ref{fig:method_overview}) as follows: 1) Object Locating, 2) Weighting Feature Map and 3) Masking Target Region.

\subsection{Object Locating with Gradient}
\label{subsec_1}
The anchor box technique is used in most detector models like Faster-RCNN\cite{ren2015faster}, YOLOX\cite{ge2021yolox}, TOOD\cite{feng2021tood}, and PAFNet\cite{xin2021pafnet} to predict the bounding boxes.
In the final feature map, each pixel predicts $N$ bounding boxes (usually $N=3$) and one bounding box for \textit{anchor free} technique.
To get the correct pixel representing the box we aim to explain, we take the derivative of the target box with the final feature map to get the location map $G_k^{l(c)}$ as the following formula:
\begin{equation} \label{gradient}
    G_k^{l(c)} = \frac{\partial S^c}{\partial A_k^l}
\end{equation}
where $G_k^{l(c)}$ denotes the gradient map of layer $l$ for feature map $k$.
$\frac{\partial S^c}{\partial A_k^l}$ is the derivative of the target class score $S^c$ with the feature map $A_k$.
In the regression task of most one-stage object detectors, $1\times1$ Convolution usually is used for predicting the bounding box, so in backward-pass, we have the Gradient map $G$ having the value of 1 pixel.
While in the two-stage object detector, because the regression and classification tasks are in two separate branches, we create a simple guide for implementing G-CAME for two-stage models in Sec.~\ref{sec:guideline}.


\subsection{Weighting feature map via Gradient-based method}
\label{subsec_2}
We adopt a gradient-based method as GradCAM~\cite{selvaraju2017grad} for classification to get the weight for each feature map. 
GradCAM method can be represented as:
\begin{equation}
    L^c_{GradCAM} = ReLU\bigg (\sum_k \alpha_k^c A_k^c \bigg )
\end{equation}
\begin{equation}
    \alpha_k^c = \sum_i \sum_j \dfrac{\partial S^c}{\partial A_{ij}^k}
\end{equation}
where $\alpha_k^c$ is the weight for each feature map $k$ of target layer $l$ calculated by taking the mean value of the gradient map $G_k^{l(c)}$.
GradCAM method produces the saliency map by linearly combining all the weighted feature maps $A_k^l$, then uses the $ReLU$ function to remove the pixel not contributing to the prediction.

As the value in gradient map can be either positive or negative, we divide all $k$ feature map into two parts ($k_1$ and $k_2$, $k_1 + k_2 = k$), the one with positive gradient $A_k^{c(+)}$ and another with negative gradient $A_k^{c(-)}$.
The negative $\alpha$ is considered to reduce the target score, so we add two parts separately and then subtract the negative part from the positive one (as Eq.~\ref{our_CAM}) to get a smoother saliency map.
\begin{equation} \label{neg_A}
    A_{k_2}^{c(-)} = \alpha_{k_2}^{c(-)}A_{k_2}^c
\end{equation}
\begin{equation} \label{pos_A}
    A_{k_1}^{c(+)} = \alpha_{k_1}^{c(+)}A_{k_1}^c
\end{equation}
\begin{equation} \label{our_CAM}
    L^c_{CAM} = ReLU \bigg (\sum_{k_1} A_{k_1}^{c(+)} - \sum_{k_2} A_{k_2}^{c(-)} \bigg )
\end{equation}

Because GradCAM can only explain classification models, it highlights all objects of the same class $c$.
By detecting the target object's location, we can guide the saliency map to only one object and make it applicable to the object detection problem.

\subsection{Masking target region with normal distribution}
\label{subsec_3}
To deal with the localization issue, we proposed to use the normal distribution to estimate the region around the object's center.
Because the gradient map shows the target object's location, we estimate the object region around the pixel representing the object's center by using a Gaussian mask as the weight for each pixel in the weighted feature map $k$. 
The Gaussian kernel is defined as:
\begin{equation} \label{gauss}
    G_\sigma = \frac{1}{2\pi\sigma^2} \exp^{-\frac{(x^2 + y^2)}{2\sigma^2}}
\end{equation}
where the term $\sigma$ is the standard deviation of the value in the Gaussian kernel and controls the kernel size.
$x$ and $y$ are two linear-space vectors filled with value in range $[1, kernel{\text -}size]$ one vertically and another horizontally. 
The bigger $\sigma$ is, the larger highlighted region we get. 
For each feature map $k$ in layer $l$, we apply the Gaussian kernel to get the region of the target object and then sum all these weighted feature maps. 
In general, we slightly adjusted the weighting feature map (Eq.~\ref{our_CAM}) to get the final saliency map as shown in Eq.~\ref{GCAM}:
\begin{equation}
\begin{aligned}
    L^c_{GCAME} = ReLU\bigg (\sum_{k_1} G_{\sigma(k_1)} \odot A_{k_1}^{c(+)} - \\ \sum_{k_2} G_{\sigma(k_2)} \odot A_{k_2}^{c(-)} \bigg )
    \label{GCAM}
\end{aligned}
\end{equation}

\subsubsection{Choosing $\sigma$ for Gaussian mask}
The Gaussian masks are applied to all feature maps, with the kernel size being the size of each feature map, and the $\sigma$ is calculated as in Eq.~\ref{sigma}.

\begin{equation} \label{R}
    R = \log \left| \frac{1}{Z} \sum_i \sum_j G_k^{l(c)}\right|
\end{equation}
\begin{equation} \label{S}
    S = \sqrt{\frac{H \times W}{h \times w}}
\end{equation}
\begin{equation} \label{sigma}
    \sigma = R\log{S} \times \frac{3}{\floor*{\frac{\sqrt{h \times w}-1}{2}}}
\end{equation}
Here, the $\sigma$ is combined by two terms. 
In the first term, we calculate the expansion factor with $R$ representing the importance of location map $G_k^{l(c)}$ and $S$ is the scale between the original image size ($H\times W$) and the feature map size ($h \times w$). We use the logarithm function to adjust the value of the first term so that its value can match the size of the gradient map.
Because object detectors usually are multi-scale object detection, we have a different $S$ for each scale level. 
For the second term, we follow the rule of thumb in choosing Gaussian kernel size as the Eq.~\ref{org_gauss} and take the inverse value.

\begin{equation} \label{org_gauss}
    kernel{\text -}size = 2 \times \ceil*{3\sigma} + 1
\end{equation}

\subsubsection{Gaussian mask generation}
We generate each Gaussian mask by following steps:
\begin{enumerate}
    \item Create a grid filled with value in range $[0, w]$ for the width and $[0, h]$ for the height ($w$ and $h$ is the size of the location map $G_k^{l(c)}$).
    \item Subtract the grid with value in position $(i_t, j_t)$ where $(i_t, j_t)$ is the center pixel of the target object on the location map.
    \item Apply Gaussian formula (Eq.~\ref{gauss}) with $\sigma$ as the expansion factor as Eq.~\ref{sigma} to get the Gaussian distribution for all values in the grid.
    \item Normalize all values in range $[0,1]$.
\end{enumerate}
By normalizing all values in range $[0,1]$, Gaussian masks only keep the region relating to the object we aim to explain and remove other unrelated regions in the weighted feature map.

%% file: results.tex
We performed our experiment on the MS-COCO 2017~\cite{lin2014microsoft} dataset with 5000 validation images.
The models in our experiment are YOLOX-l (one-stage model) and Faster-RCNN (two-stage model). 
All experiments are implemented in Pytorch~\cite{paszke2019pytorch} and conducted on NVIDIA Tesla P100 GPU.
G-CAME's inference time depends on the number of feature maps in selected layer $l$.
Our experiments run on model YOLOX-l with 256 feature maps for roughly $0.5s$ per object.

\subsection{Saliency map visualization}
\begin{figure} [ht!]
    \centering
    \includegraphics[scale=0.455]{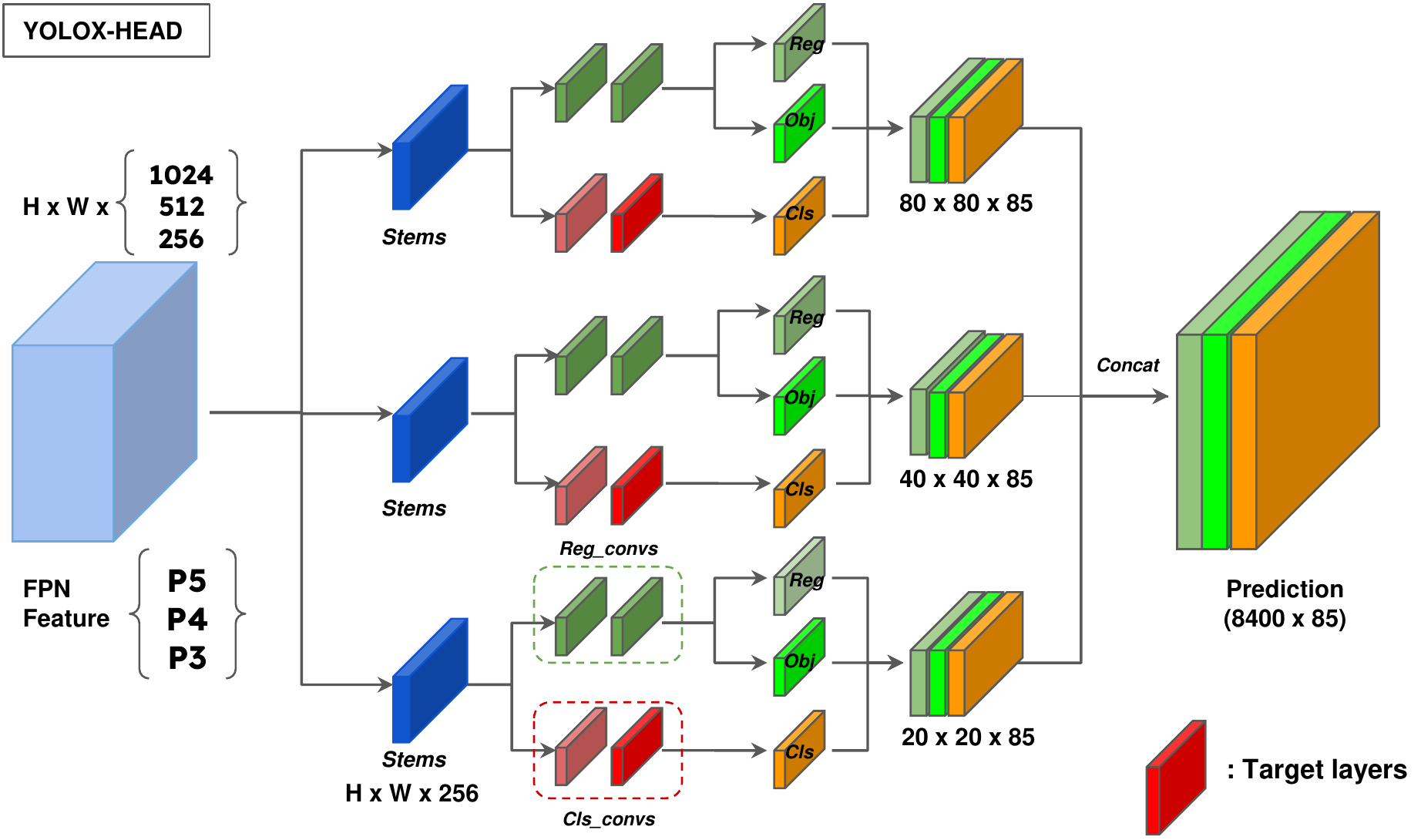}
    \caption{We choose the last convolution layer in predicting the class phase of each branch of YOLOX-l-HEAD as the target layer for G-CAME}.
    \label{fig:experiment}
\end{figure}

We performed a saliency map qualitative comparison of G-CAME with D-RISE~\cite{petsiuk2021black} to validate the results of G-CAME.
We use D-RISE's default parameters~\cite{petsiuk2021black}, where each grid's size is $16\times16$, the probability of each grid's occurrence is $0.5$, and the amount of samples for each image is $4000$.
For G-CAME, we choose the final convolution layer in each branch of YOLOX as the target layer to calculate the derivative (Fig.~\ref{fig:experiment}).

\begin{figure*}[tbh!]
    \centering
    \includegraphics[]{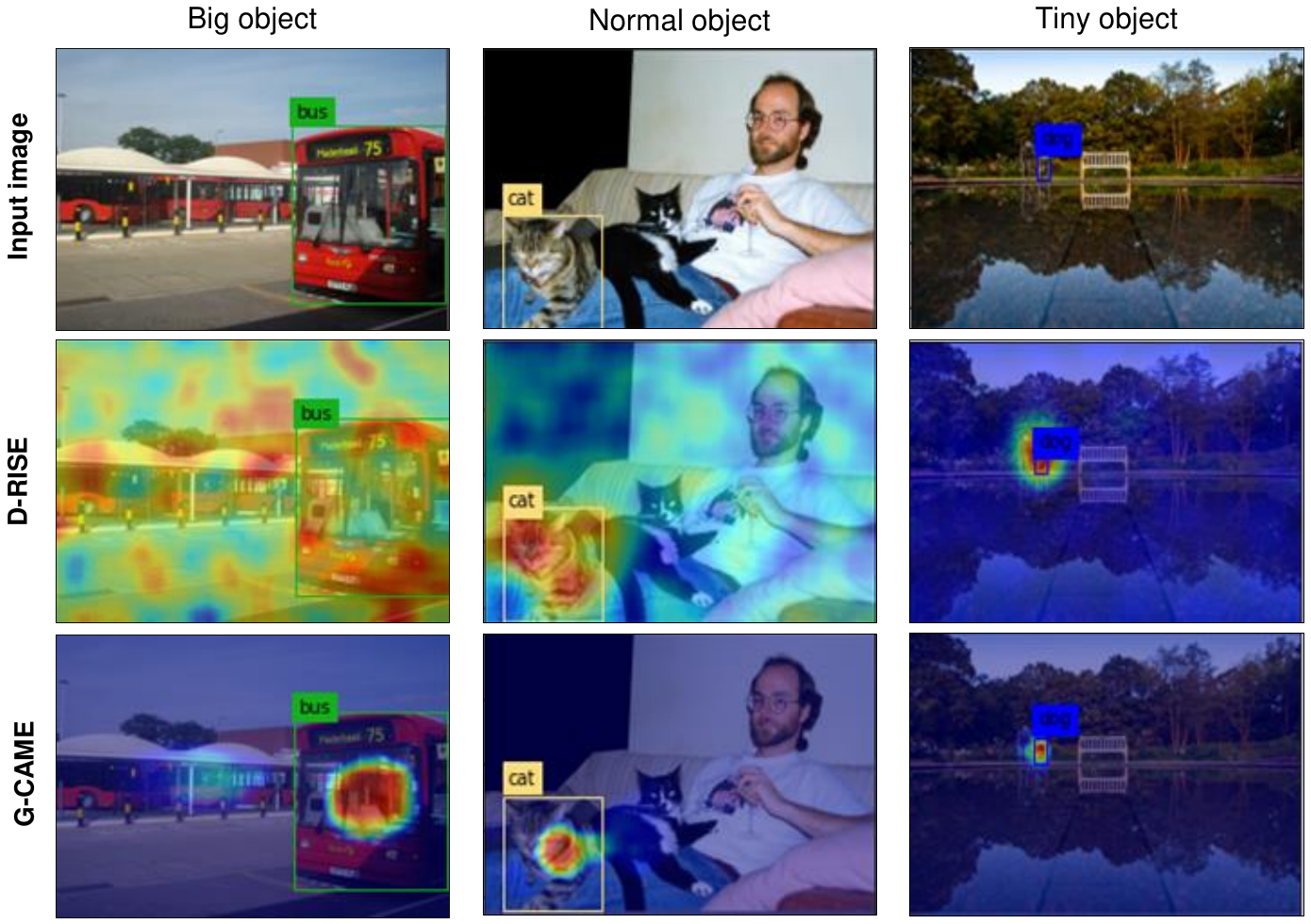}
    \caption{Visualization results of D-RISE and G-CAME on three sizes of the object: big object, normal object, and tiny object. In all cases, results show that G-CAME's saliency maps are more precise and have less noise than D-RISE's.}
    \label{fig:compare_rs}
\end{figure*}

Fig.~\ref{fig:compare_rs} shows the results of G-CAME compared with D-RISE.
As can be seen from the result, G-CAME significantly reduced random noises.
Also, G-CAME can generate smoother saliency maps compared with D-RISE in a short time.

\subsection{Localization Evaluation}
To evaluate the new method, we used two standard metrics, Pointing Game~\cite{zhang2018top} and Energy-based Pointing Game~\cite{wang2020score} to compare the correlation between an object's saliency map and human-labeled ground truth. 
The results are shown in Table~\ref{tab:rs_comparison}.

\subsubsection{Pointing Game (PG)}
We used the pointing game metric~\cite{zhang2018top} as a human evaluation metric.
Firstly, we run the model on the dataset and get the bounding boxes that best match the ground truth for each class on each image.
A $hit$ is scored if the highest point of the saliency map lies inside the ground truth; otherwise, a $miss$ is counted. 
The pointing game score is calculated by $PG = \frac{{\#}Hits}{{\#}Hits + {\#}Misses}$ for each image. 
This score should be high for a good explanation to evaluate an XAI method.

\subsubsection{Energy-Based Pointing Game (EBPG)}
EBPG~\cite{wang2020score} calculates how much the energy of the saliency map falls inside the bounding box.
EBPG formula is defined as follows:
\begin{equation} \label{ebpg}
    Proportion = \frac{\sum L^c_{(i, j) \in bbox}}{L^c_{(i, j) \in bbox} + L^c_{(i, j) \notin bbox}}
\end{equation}
Similar to the PG score, a good explanation is considered to have a higher EBPG.

PG and EBPG results are reported in Table \ref{tab:rs_comparison}.
Specifically, more than 65\% energy of G-CAME's saliency map falls into the ground truth bounding box compared with only 18.4\% of D-RISE.
In other words, G-CAME drastically reduces noises in the saliency map. 
In Pointing game evaluation, G-CAME also gives better results than D-RISE. 98\% of the highest pixel lie inside the correct bounding box, while this number in D-RISE is 86\%.

\begin{table}[tbh!]
\centering
\begin{tabular}{ccc}
\toprule
Method & D-RISE    & G-CAME (Our)      \\ 
\midrule
\begin{tabular}[c]{@{}c@{}}PG\\\textit{(Overall \textbar{} Tiny object)}\end{tabular}               & 0.86 \textbar{} 0.127  & \textbf{0.98 \textbar{} 0.158}   \\ 
\midrule
\begin{tabular}[c]{@{}c@{}}EBPG\\\textit{(Overall \textbar{} Tiny object)}\end{tabular} & 0.184 \textbar{} 0.009 & \textbf{0.671 \textbar{} 0.261}  \\
\bottomrule
\end{tabular}
\caption{\label{tab:rs_comparison}Localization evaluation between D-RISE and G-CAME on the MS-COCO 2017 validation dataset. The best is in bold.}
\end{table}

\subsubsection{Bias in Tiny Object Detection}
\begin{figure*}
    \centering
    \begin{subfigure}[b]{0.505\textwidth}
        \centering
        \includegraphics[width=\textwidth]{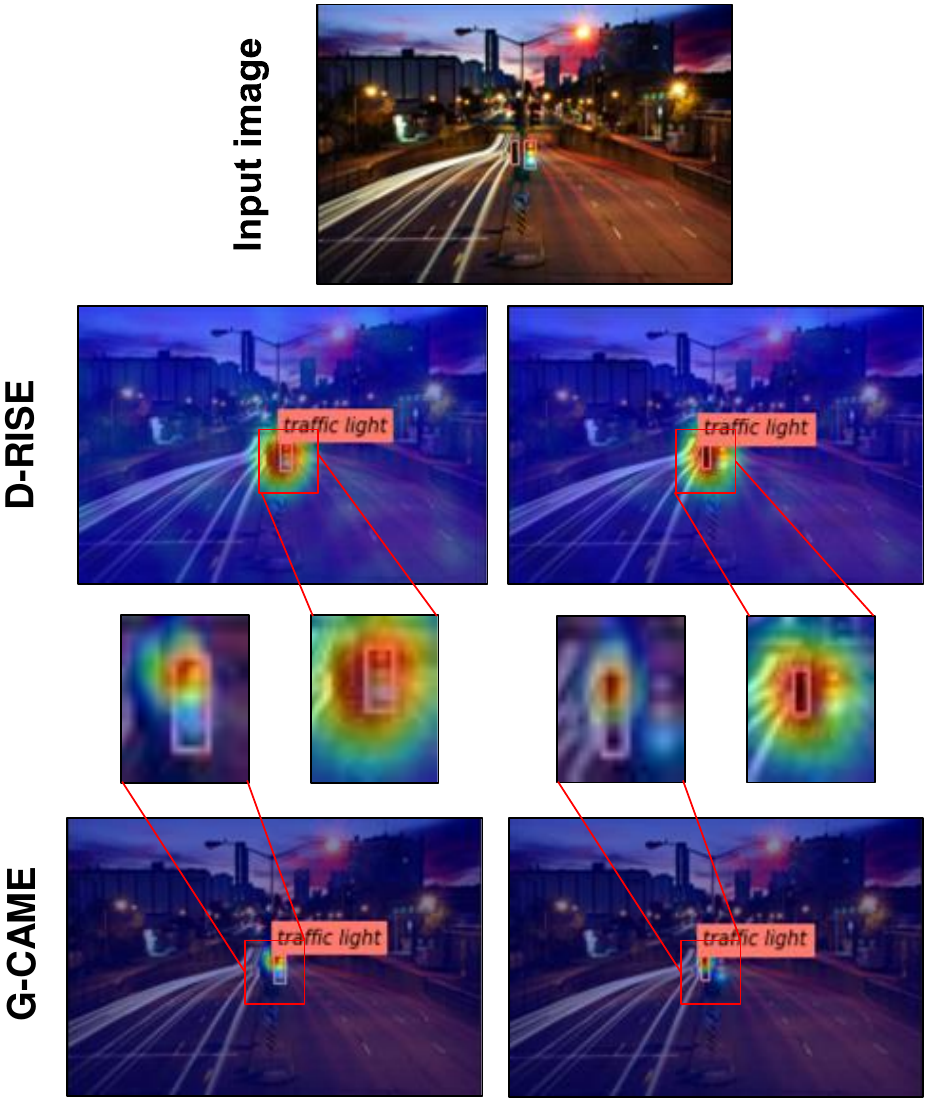}
        \caption{}
        \label{subfig:tiny1}
    \end{subfigure}
    \hfill
    \begin{subfigure}[b]{0.48\textwidth}
        \centering
        \includegraphics[width=\textwidth]{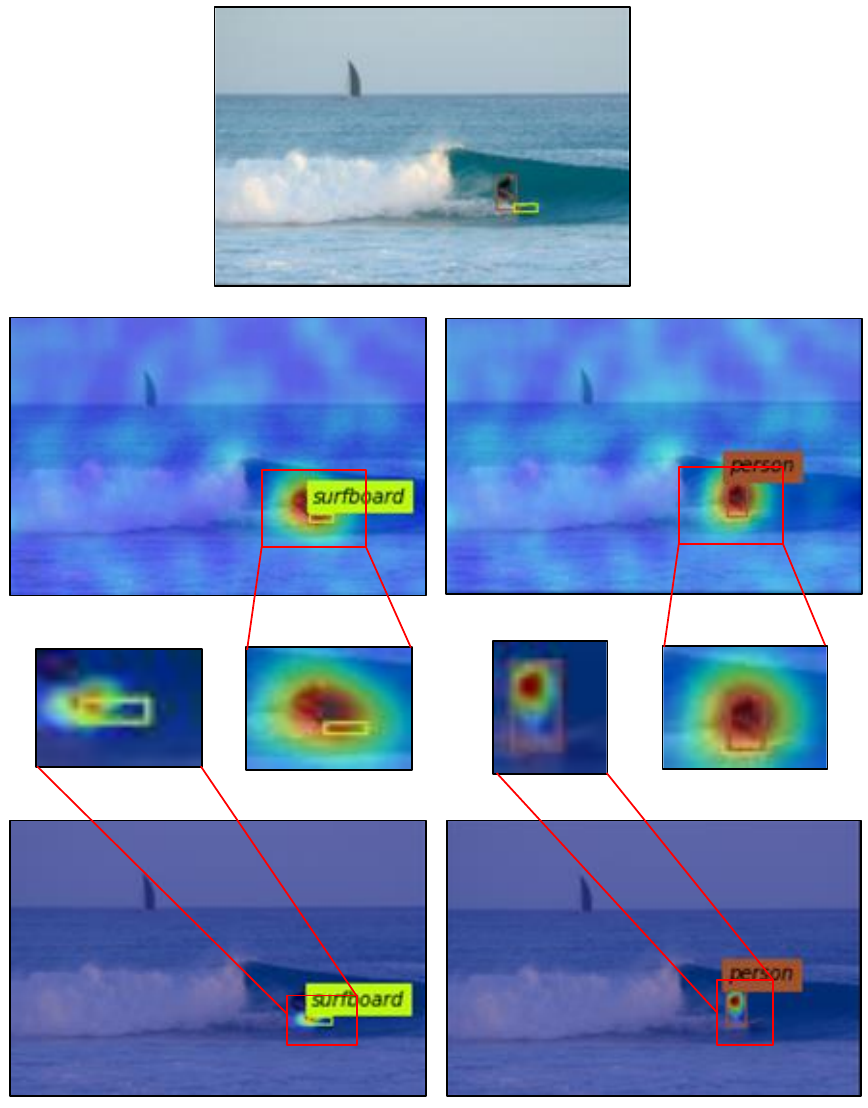}
        \caption{}
        \label{subfig:tiny2}
    \end{subfigure}
    \caption{The saliency map of D-RISE and G-CAME for tiny object prediction. We test two methods in two cases: tiny objects of the same class lying close together (Fig.~\ref{subfig:tiny1}) and multiple objects of different classes lying close together (Fig.~\ref{subfig:tiny2}). In both cases, G-CAME can clearly distinguish each object in its explanation.}
    \label{fig:tiny_obj}
\end{figure*}

Explaining tiny objects detected by the model can be a challenge.
In particular, the saliency map may bias toward the neighboring region. 
This issue can worsen when multiple tiny objects partially or fully overlap because the saliency map  stays in the same location for every object. 
In our experiments, we define the tiny object by calculating the ratio of the predicted bounding box area to the input image area (640$\times$640 in YOLOX).
An object is considered tiny when this ratio is less than or equal to 0.005.
In Fig.~\ref{fig:tiny_obj}, we compare our method with D-RISE in explaining tiny object prediction for two cases. 
In the first case (Fig.~\ref{subfig:tiny1}), we test the performance of D-RISE and G-CAME in explaining two tiny objects of the same class.
The result shows that D-RISE fails to distinguish two ``traffic lights'', where the saliency maps are nearly identical.
For the case of multiple objects with different classes overlapping (Fig.~\ref{subfig:tiny2}), the saliency maps produced by D-RISE hardly focus on one specific target.
The saliency corresponding to the ``surfboard'' even covers the ``person'', and so does the explanation of the ``person''.
The problem can be the grid's size in D-RISE, but changing to a much smaller grid's size can make the detector unable to predict. 
In contrast, G-CAME can clearly show the target object's localization in both cases and reduce the saliency map's bias to unrelated regions. 
In detail, we evaluated our method only in explaining tiny object prediction with EBPG score.
The MS-COCO 2017 validation dataset has more than 8000 tiny objects, and the results are reported in Table~\ref{tab:rs_comparison}. 
Our method outperforms D-RISE with more than 26\% energy of the saliency map falling into the predicted box, while this figure in D-RISE is only 0.9\%. 
Especially, most of the energy in D-RISE's explanation does not focus on the correct target.
In the PG score, instead of evaluating one pixel, we assess all pixels having the same value as the pixel with the highest value.
The result also shows that G-CAME's explanation has better accuracy than D-RISE's.

\subsection{Faithfulness Evaluation}
A good saliency map for one target object should highlight regions that affect the model's decision most.
So, we employ the \textit{Average Drop (AD)} metric to evaluate the confidence change~\cite{chattopadhay2018grad, fu2020axiom, ramaswamy2020ablation} in the model's prediction for the target object when using the explanation as the input.
In other words, when we remove these important regions, the confidence score of the target box should be reduced.
The \textit{Average Drop} can be calculated by the formula:
\begin{equation} \label{Avg_drop}
    AD = \frac{1}{N}\sum_{i=1}^N \frac{max(P_c(I_i) - P_c(\tilde{I_i}), 0)}{P_c(I_i)} \times 100
\end{equation}
where:
\begin{equation} 
    \tilde{I_o} = I \odot (1 - M_o) + \mu M_o
    \label{perturbed}
\end{equation}
\begin{equation} 
    P_c(\tilde{I}) = IOU(L_i, L_j) \cdot p_{c(L_j)}
    \label{prob}
\end{equation}

Here, we adjust the original formula of \textit{Average Drop} for the object detection model.
In Eq.~\ref{perturbed}, we create a new input image masked by the explanation $M$ of G-CAME.
$\mu$ is the mean value of the original image.
With the value of $M$, we only keep 20\% of the pixel with the most significant value in the original explanation and set the rest as 0.
Then, we can minimize the explanation's noise, and the saliency map can focus on the regions most influencing the prediction. 
 
In Eq.~\ref{prob}, to compute probability $P_c(\tilde{I})$, we first calculate the pair-wise $IOU$ of the box $L_j$ predicted on perturbed image $\tilde{I}$ with the box $L_i$ predicted on the original image and take the one with the highest value.
After that, we multiply the first term with the corresponding class score $p_{c(L_j)}$ of the box.
In calculating $P_c(I_i)$, the $IOU$ equals 1, so the value remains the original confidence score. 
Hence, if the explanation is faithful, the confidence drop should increase.

However, removing several pixels can penalize the method of producing the saliency map that has connected and coherent regions.
Specifically, pixels representing the object's edges are more meaningful than others in the middle~\cite{kapishnikov2019xrai}.
For example, pixels representing the dog's tail are easier to recognize than others lying on the dog's body.

To give a comparison when using the confidence drop score, we compare the \textit{information level} of the \textit{bokeh} image, which is created by removing several pixels from the original image, after applying the XAI method.
To measure the \textit{bokeh} image's information, we use WebP~\cite{Webp} format and calculate the drop information by taking the proportion of the compressed size of the \textit{bokeh} image to the original image~\cite{kapishnikov2019xrai}.

Table~\ref{tab:avg_drop_rs} shows the confidence and information drop results.
In detail, D-RISE performs better in the drop confidence score, with a 42.3\% reduction in the predicted class score when removing the highest value pixel.
In the drop information score, our method achieves 29.1\% compared to 31.58\% of D-RISE, which means that our method preserves the original image's information better than D-RISE.
Moreover, since G-CAME inherits the CAM-based strength in running time, G-CAME takes under 1 second to explain, while D-RISE needs roughly 4 minutes to run on the same benchmark.
Because of employing feature maps as a part of the explanation, G-CAME can also reflect what the model focuses on predicting, while D-RISE cannot.

\begin{table}[h]
\centering
\resizebox{.85\linewidth}{!}{%
\begin{tabular}{ccc}
\toprule
Method          & D-RISE & G-CAME (Our) \\
\midrule
Confidence Drop\%$\uparrow$ & \textbf{42.3} & 36.8 \\
\midrule
Information Drop\%$\downarrow$ & 31.58 & \textbf{29.15} \\
\midrule
Running time(s)$\downarrow$ & 252 & \textbf{0.435} \\
\bottomrule
\end{tabular}}
\caption{\label{tab:avg_drop_rs}Confidence drop score, Information drop score, and Running time of D-RISE and G-CAME with YOLOX-l on MS-COCO 2017 dataset. The arrows $\uparrow/\downarrow$ indicate higher or lower scores are better. The best is shown in bold.}
\end{table}

\subsection{Sanity check}
\begin{figure} [tbh!]
    \centering
    \includegraphics[scale=0.56]{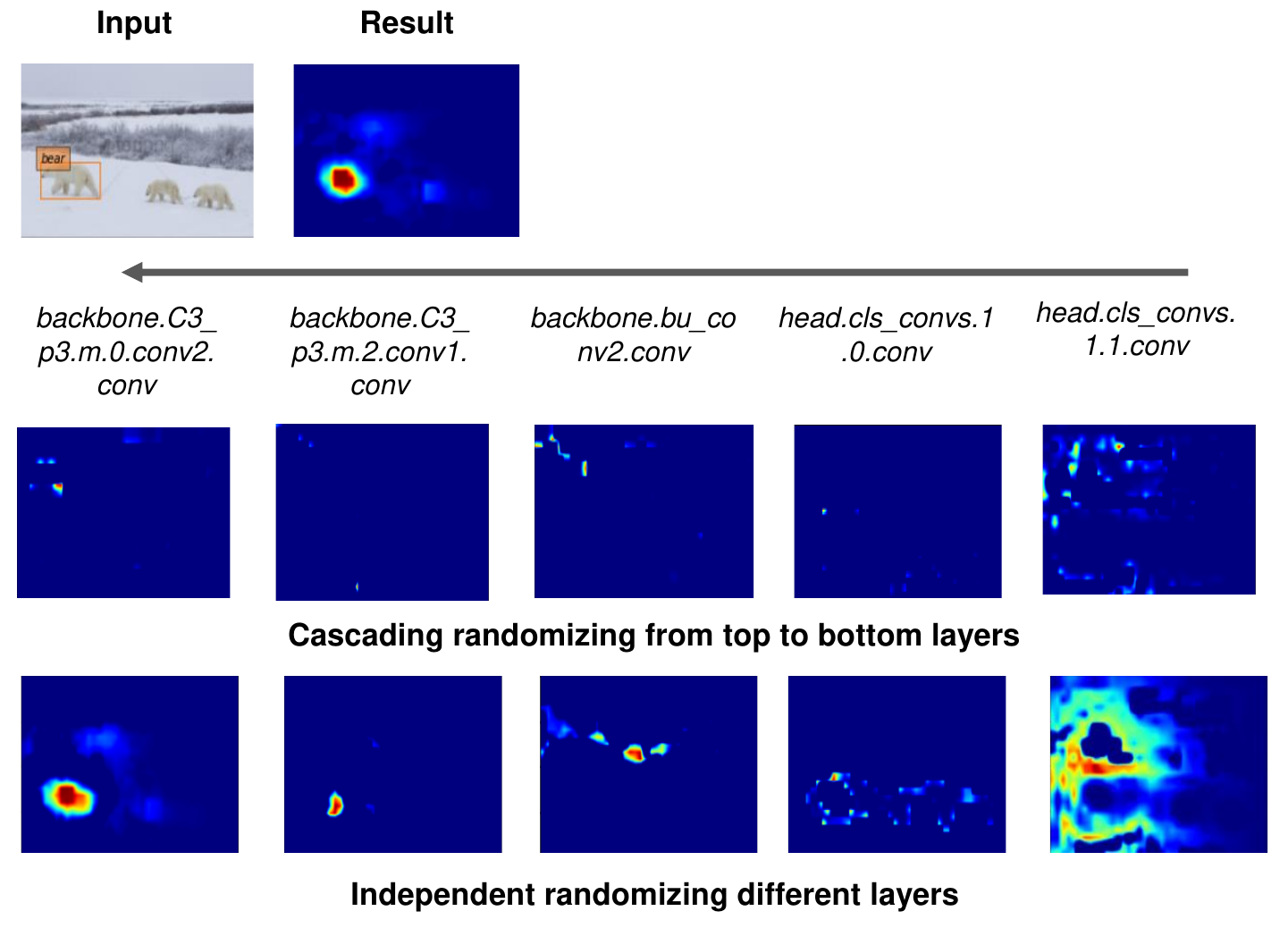}
    \caption{The result of \textit{Cascading Randomization} and \textit{Independent Randomization} for five layers from top to bottom of the YOLOX model. Chosen layers in the head part do not include the layer in the regression branch. The result shows our method is sensitive to the model's parameters}.
    \label{fig:sanity_check}
\end{figure}

To validate whether the saliency map is a faithful explanation or not, we perform a sanity check~\cite{adebayo2018sanity} with \textit{Cascading Randomization} and \textit{Independent Randomization}. 
In \textit{Cascading Randomization} approach, we randomly choose five convolution layers as the test layers.
Then, for each layer between the selected layer and the top layer, we remove the pre-trained weights, reinitialize with normal distribution, and perform G-CAME to get the explanation for the target object.
In contrast to \textit{Independent Randomization}, we only reinitialize the weight of the selected layer and retain other pre-trained weights.
The sanity check results show that G-CAME is sensitive to model parameters and can produce valid results, as shown in Fig.~\ref{fig:sanity_check}. 

\subsection{Approach for two-stage model (Faster-RCNN)}
\label{sec:guideline}
This section extends G-CAME's application to a two-stage model, namely Faster-RCNN~\cite{ren2015faster}.
In Faster-RCNN, the image is first passed through several stacked convolution layers to extract features. 
The Region Proposal Network (RPN) detects regions possibly containing an object.
Those regions are then fed to the Region of Interest (ROI) Pooling layer to be in a fixed size. 
After that, two 1$\times 1$ convolution layers, including a classification layer to detect the probability of an object's occurrence and a regression layer to detect the coordinate of bounding boxes, are used to detect the bounding boxes.
The output bounding boxes are passed through the Faster-RCNN Predictors, including two fully connected layers.
Then, G-CAME utilizes the feature maps at the end of the feature extraction phase to explain. 

First, we calculate the partial derivative of the class score according to each feature map of selected layers.
Faster-RCNN has four branches of detecting objects, and we choose the last convolution layer of each branch to calculate the derivative.
When we take the derivative of the class score to the target layer, the gradient map ($G_k^{l(c)}$) has more than one pixel having value because anchor boxes are created in the next phase, namely the detecting phase.
Thus, we cannot get the pixel representing the object's center through the gradient map.
To solve this issue, we set the pixel with the highest value in the gradient map as the center of the Gaussian mask.
We estimate that the area around the highest value pixel likely contains relevant features.
We perform the same in the \textit{Weighting feature map} and \textit{Masking region} phases as in Fig.~\ref{fig:method_overview}.

%% file: conclusion.tex
In this paper, we proposed G-CAME, a new method to explain object detection models motivated by the CAM-based method and Gaussian kernel.
A simple guide is provided to implement our method in both one-stage and two-stage detectors.
The experiment results show that our method can plausibly and faithfully explain the model's predictions.
Moreover, our method runs reasonably short, which overcomes the time constraint of existing perturbation-based methods and reduces the noise in the saliency map.

%% file: appendix.tex
This supplementary material provides our experiment of applying G-CAME on the Faster-RCNN~\cite{ren2015faster} model and visualization comparison between G-CAME and D-RISE~\cite{petsiuk2021black}.
We also clarify how we evaluate our method with the drop confidence score.
Finally, we provide more sanity check results with different layers in YOLOX-l~\cite{ge2021yolox}.

\section{G-CAME for the Faster-RCNN model}
\begin{figure*}[t!]
    \centering
    \includegraphics[scale=0.7]{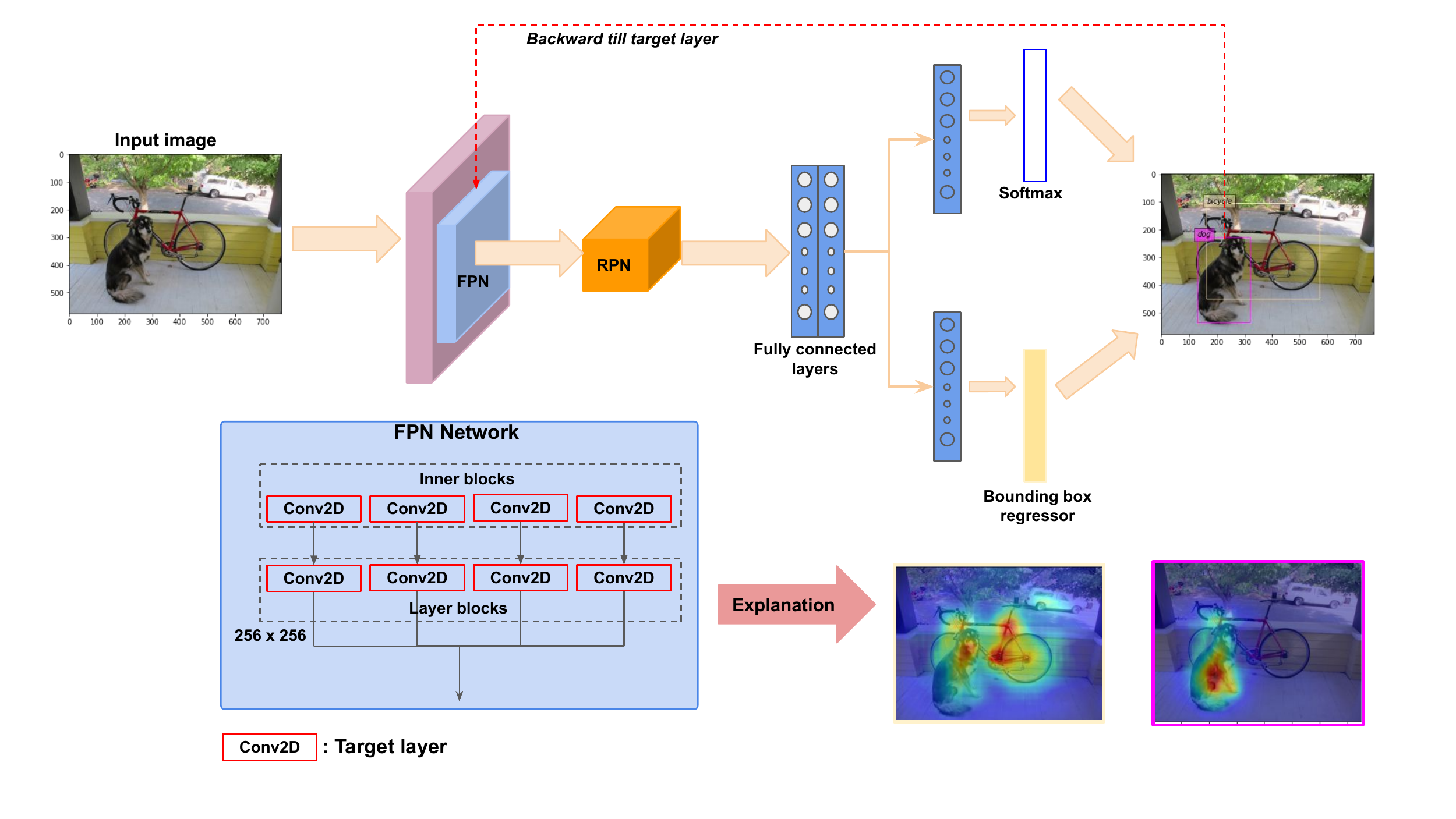}
    \caption{To apply G-CAME on Faster-RCNN, we choose the convolution layers in the FPN network as the target layers to perform backward.}
    \label{fig:faster_pipeline}
\end{figure*}

In Fig.~\ref{fig:faster_pipeline}, we introduce a guide for choosing the target layers in Faster-RCNN to apply G-CAME.
In Faster-RCNN, features are extracted in backbone layers and passed through the Feature Pyramid Network (FPN)~\cite{lin2017feature} network, which includes four branches to detect the different objects' sizes.
Hence, we choose the convolution layers in the FPN network as the target layers to analyze.

\subsection{Visualize saliency map}
\begin{figure*}[]
    \centering
    \begin{tabular}{@{}c@{}}
    \includegraphics[scale=0.78]{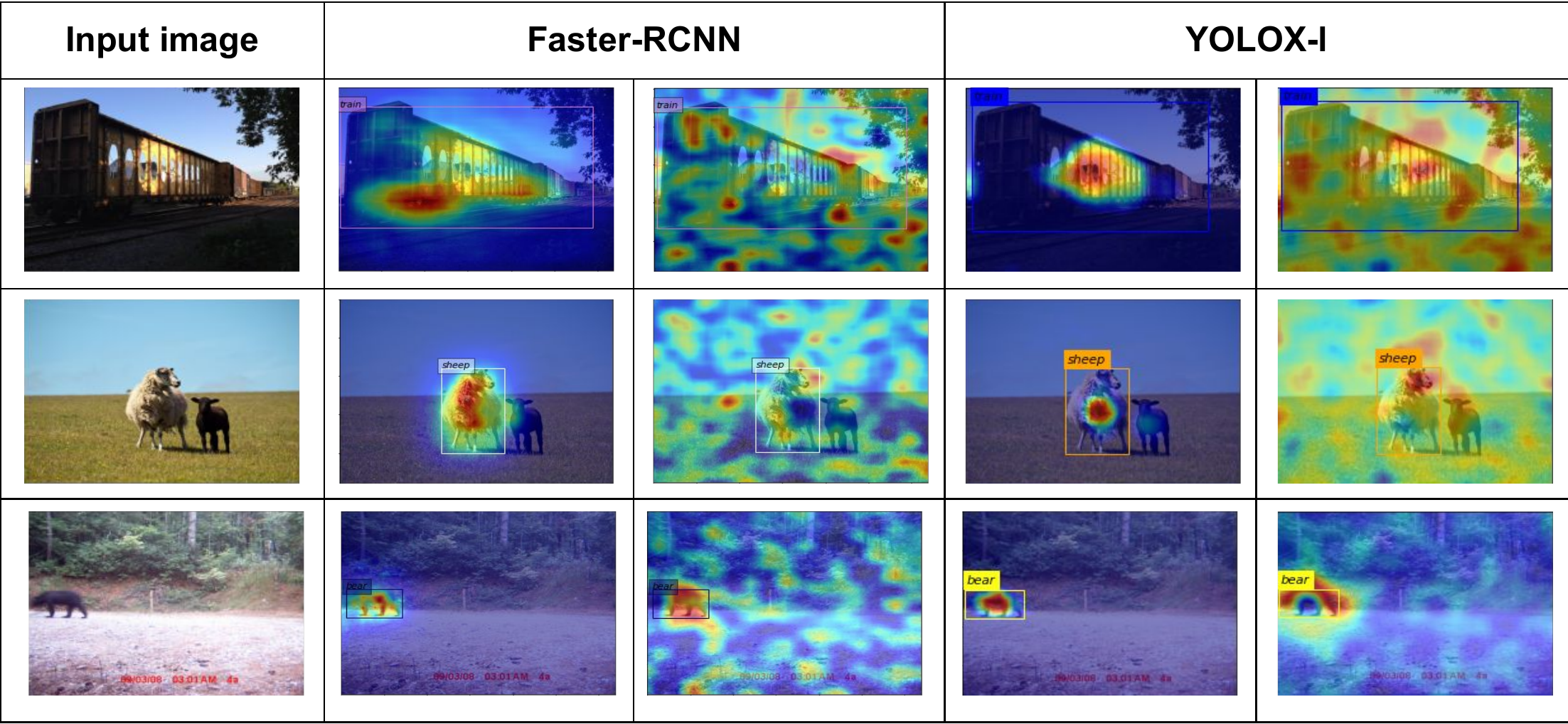} \\
    \includegraphics[scale=0.78]{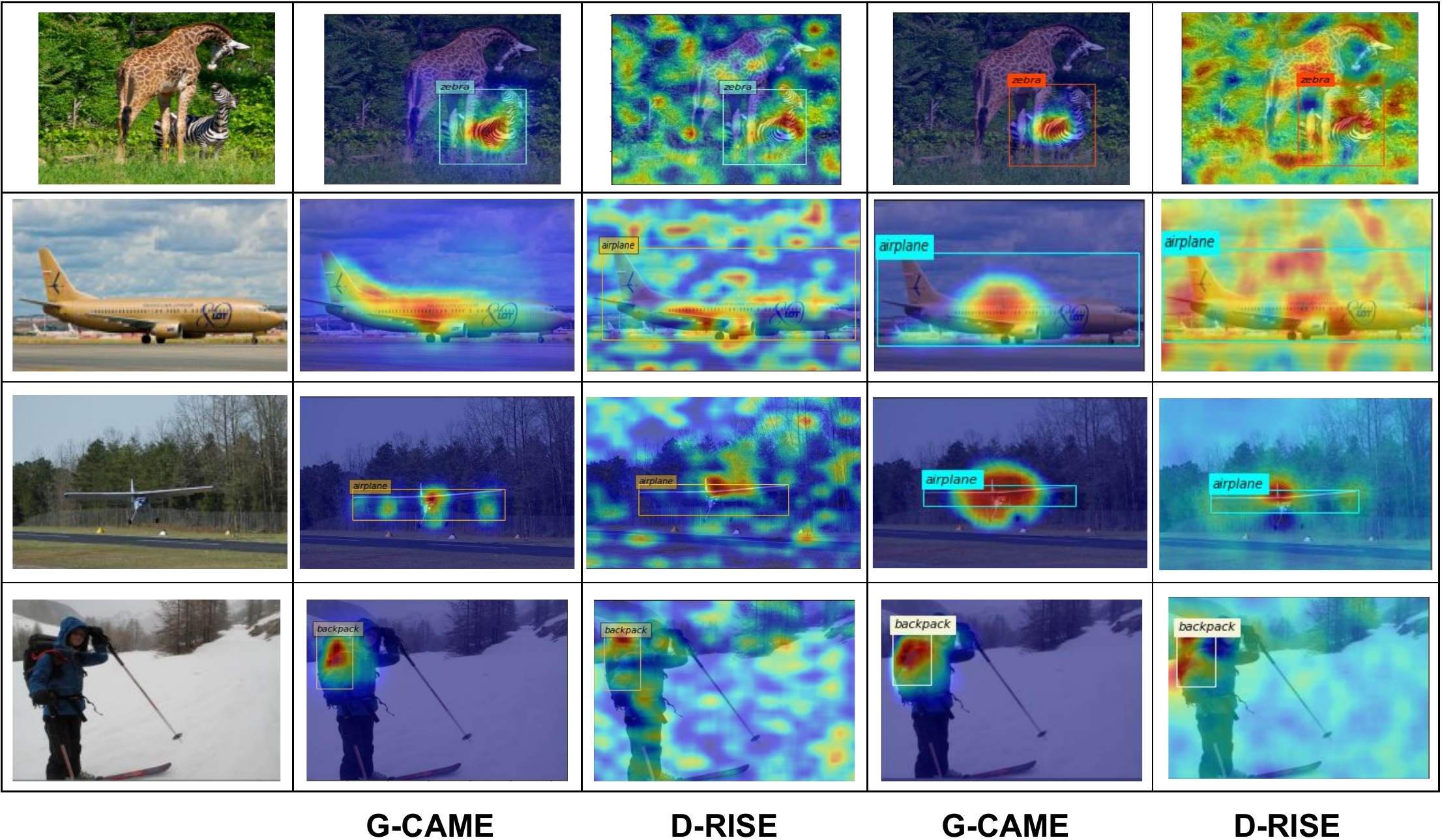}
    \end{tabular}
    \caption{Visualization comparison between G-CAME and D-RISE when applying two models, FasterRCNN and YOLOX-l.}
    \label{fig:comparison}
\end{figure*}

In this section, we provide more visualization results of G-CAME compared with D-RISE~\cite{petsiuk2021black}, which are shown in Fig.~\ref{fig:comparison}.

\subsection{Faithfulness Evaluation}
\begin{figure}[h]
    \centering
    \includegraphics[scale=0.5]{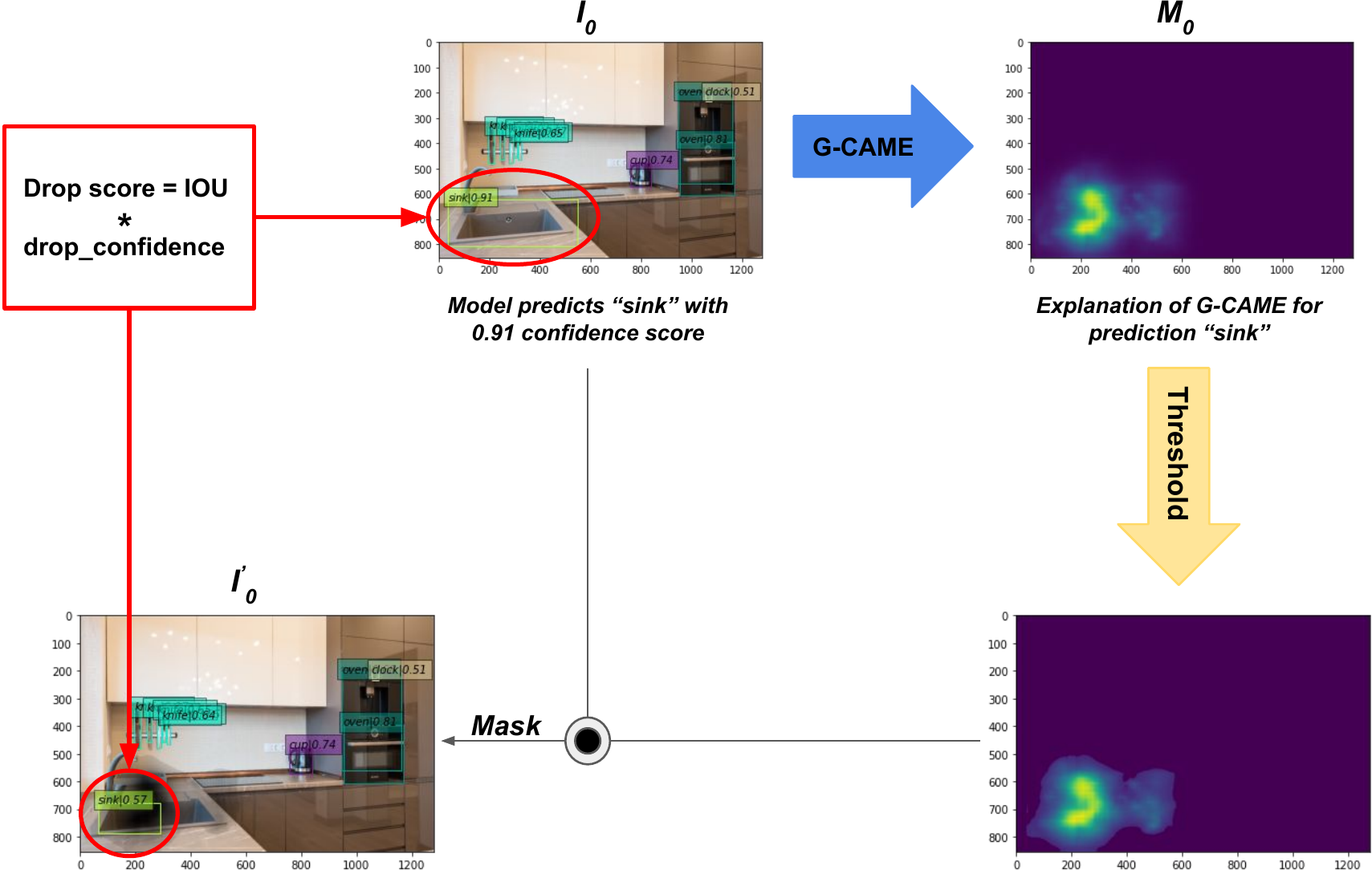}
    \caption{How to calculate the Drop Confidence score}
    \label{fig:drop_conf}
\end{figure}

This section illustrates how to evaluate an XAI method with drop confidence score~\cite{fu2020axiom}.
In our experiment, the average drop confidence is calculated as follows:
\begin{equation} \label{Avg_drop}
    AD = \frac{1}{N}\sum_{i=1}^N \frac{max(P_c(I_i) - P_c(\tilde{I_i}), 0)}{P_c(I_i)} \times 100
\end{equation}
where:
\begin{equation} 
    \tilde{I_o} = I \odot (1 - M_o) + \mu M_o
    \label{perturbed}
\end{equation}
\begin{equation} 
    P_c(\tilde{I}) = IOU(L_i, L_j) \cdot p_{c(L_j)}
    \label{prob}
\end{equation}
For more details, Fig.~\ref{fig:drop_conf} shows the evaluation process of explaining G-CAME.
A threshold is employed to keep 20\% of the pixel with the most value in the explanation. Then we mask the explanation with the original image by Eq.~\ref{perturbed}. 
Finally, the drop confidence score will be calculated as the formula in the red box.

\subsection{Sanity check}
\begin{figure}[h!]
    \centering
    \includegraphics[scale=0.6]{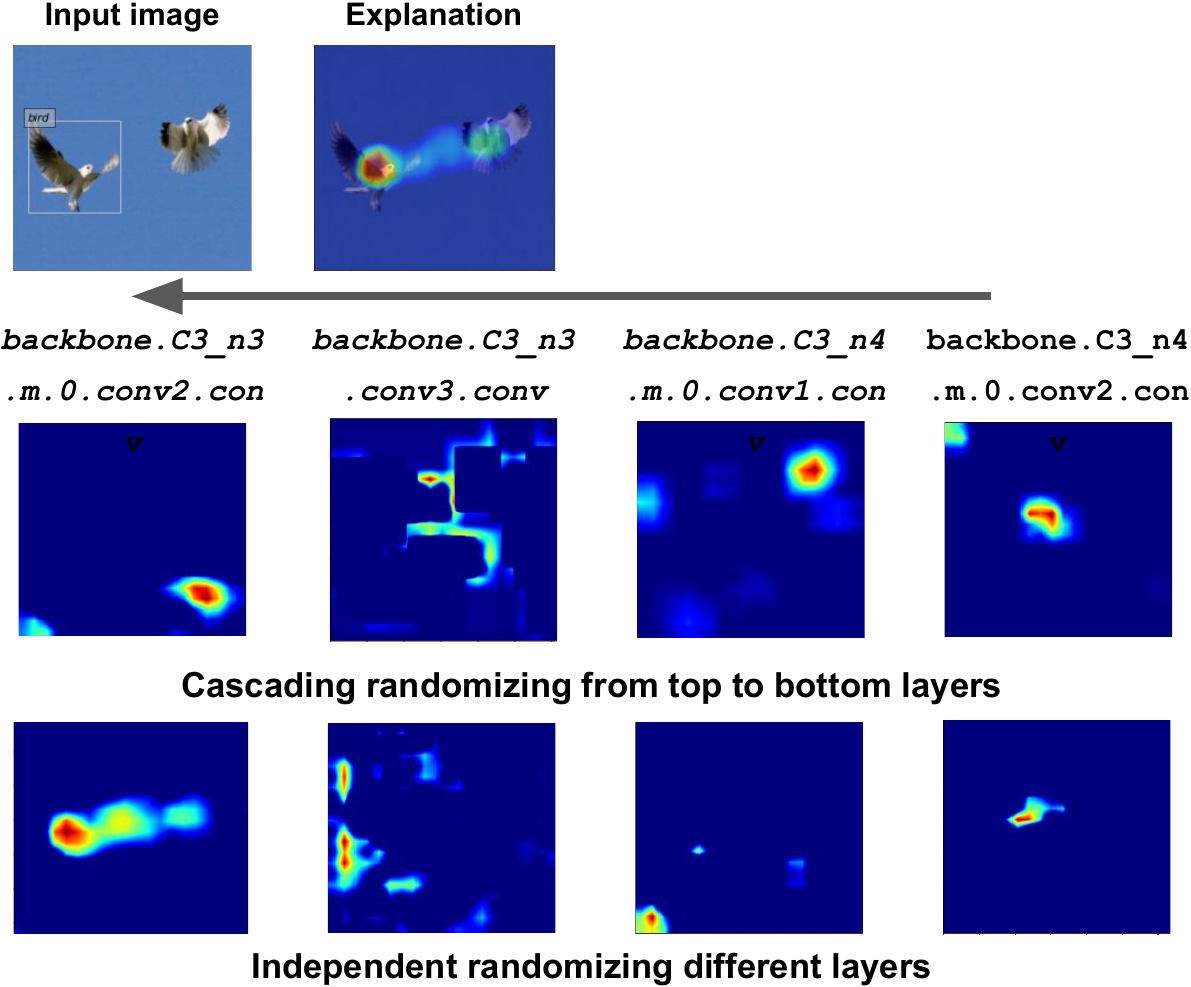}
    \caption{Results of sanity check with YOLOX-l's parameters}
    \label{fig:sup_sanity}
\end{figure}

A good XAI method has a good explanation and must be faithfully sensitive to the model's parameters.
In this section, we provide additional results of G-CAME with sanity check~\cite{adebayo2018sanity} in Fig.~\ref{fig:sup_sanity}.

